\documentclass{article} 
\usepackage{nips12submit_e,times}
\usepackage{amsmath}
\usepackage{amsthm}
\usepackage{amssymb}
\usepackage{graphicx}
\usepackage{algorithm}
\usepackage{algorithmic}

\DeclareSymbolFont{lettersA}{U}{txmia}{m}{it}
\DeclareMathSymbol{\Z}{\mathord}{lettersA}{'232} 
\DeclareMathSymbol{\R}{\mathord}{lettersA}{'222}

\usepackage{braket}

\title{Learning dynamic Boltzmann machines with spike-timing dependent plasticity}

\author{Takayuki Osogami\\
IBM Research - Tokyo
\And 
Makoto Otsuka\\
IBM Research - Tokyo
}

\nipsfinalcopy 

\begin{document}

\thispagestyle{empty}

\begin{center} \hspace*{1cm} \end{center} 

{\tt{\flushright
September 28, 2015
 \\}
\noindent
RT0967
\\
\tt Mathematics $\:\:$ 13 pages}

\vspace {0.5cm}

\noindent{\bf {\Huge Research Report}}

\vspace {1cm}

\noindent{\Large
Learning dynamic Boltzmann machines with spike-timing dependent plasticity
}

\vspace {0.8cm}

\noindent{\Large
Takayuki Osogami and Makoto Otsuka
}

\vspace {0.8cm}

\noindent IBM Research - Tokyo \\
IBM Japan, Ltd. \\
19-21 Hakozaki, Chuo-ku, Tokyo 103-8510, Japan

\vfill

\newpage

\maketitle

\begin{abstract}
We propose a particularly structured Boltzmann machine, which we refer
to as a dynamic Boltzmann machine (DyBM), as a stochastic model of a
multi-dimensional time-series.  The DyBM can have infinitely many
layers of units but allows exact and efficient inference and learning
when its parameters have a proposed structure.  This proposed
structure is motivated by postulates and observations, from biological
neural networks, that the synaptic weight is strengthened or weakened,
depending on the timing of spikes (i.e., spike-timing dependent
plasticity or STDP).  We show that the learning rule of updating the
parameters of the DyBM in the direction of maximizing the
likelihood of given time-series can be interpreted as STDP with
long term potentiation and long term depression.  The learning rule
has a guarantee of convergence and can be performed in a distributed
matter (i.e., local in space) with limited memory (i.e., local in
time).
\end{abstract}

\section{Introduction}

Boltzmann machines have seen successful applications in recognition of
images and other tasks of machine learning \cite{TPV13, SZLL13, GN13, DAL12}, particularly with recent
development of deep learning \cite{hinton2006reducing}.  The standard
approaches to training a Boltzmann machine iteratively apply a Hebbian
rule \cite{hebb1949organization} either exactly or approximately,
where the values of the parameters are updated in the directions of
increasing the likelihood of given training data with respect to the
{\em equilibrium} distribution of the Boltzmann machine
\cite{hinton1983optimal}.  This Hebbian rule for the Boltzmann machine
is limited in the sense that the concept of time is missing.  For
biological neural networks, spike-timing dependent plasticity (STDP)
has been postulated and supported empirically
\cite{markram1997regulation,bi1998synaptic,sjostrom2001rate}.  For
example, the synaptic weight is strengthened when a post-synaptic
neuron fires shortly after a pre-synaptic neuron fires (i.e., long
term potentiation) but is weakened if this order of firing is reversed
(i.e., long term depression).

In this paper, we study the dynamics of a Boltzmann machine, or a
dynamic Boltzmann machine (DyBM)\footnote{The natural acronym, DBM, is
  reserved for Deep Boltzmann Machines \cite{SalHinton07}.}, and derive
a learning rule for the DyBM that can be interpreted as STDP.
While the conventional Boltzmann machine is trained with a collection
of static patterns (such as images), the DyBM is trained with a
time-series of patterns.  In particular, the DyBM gives the
conditional probability of the next values (patterns) of a time-series
given its historical values.  This conditional probability can depend
on the whole history of the time-series, and the DyBM can thus be used
iteratively as a generative model of a time-series.

Specifically, we define the DyBM as a Boltzmann machine having
multiple layers of units, where one layer represents the most recent
values of a time-series, and the remaining layers represent the
historical values of the time-series.  We assume that the most recent
values are conditionally independent of each other given the
historical values.  The DyBM allows an infinite number of layers, so
that the most recent values can depend on the whole history of the
time series.  We train the DyBM in such a way that the likelihood of
given time-series is maximized with respect to the conditional
distribution of the next values given the historical values.  This
definition of the DyBM and the general approach to training the DyBM
constitute the first contribution of this paper.

We show that the learning rule for the DyBM is significantly
simplified 
and exhibits various characteristics of STDP that have been observed in biological neural networks \cite{abbott2000synaptic},
when
the DyBM has an infinite number of layers
and particularly structured parameters.  Specifically, we
assume that the weight between a unit representing a most recent value
(time 0) and a unit representing the value in the past (time $-t$) is
the sum of geometric functions with respect to $t$.  We show that
updating parameters associated with a pair of units
requires only the information that is available at those units (i.e., local
in space), and the required information can be maintained by keeping a
first-in-first-out (FIFO) queue of the last values of a unit (i.e., local in time).
The convergence of the learning rule is guaranteed with
sufficiently low learning rate, because the parameters
are always updated such that the likelihood of given training data is increased.
The learning rule that is formally derived for the DyBM
and its interpretation as STDP constitute the second contribution
of this paper.

The prior work has extended the Boltzmann machine to incorporate the
timing of spikes in various ways
\cite{hinton1999spiking, sutskever2007learning, sutskever2008recurrent, taylor2009factored,
MKSL14}.
However, the existing learning rules for those extended Boltzmann
machines involve approximation with contrastive divergence and do not
have some of the characteristics of the STDP (e.g., long term
depression) that we show for the DyBM.

We will see that the DyBM can be considered as a recurrent neural
network (RNN) equipped with memory units.  The DyBM is thus
related to Long Short Term Memory \cite{LSTM, LSTMonline, LSTMoffline, GJ14}
and other RNNs
\cite{deepRNN, RNNwithSTDP, PC14, hessianfreeRNN,SMH11}.  What distinguishes the DyBM from existing
RNNs is that training a DyBM does not require ``backpropagation through
time,'' or a chain rule of derivatives.  This distinguishing feature of
the DyBM follows from the fact the DyBM can be equivalently interpreted
both as an RNN and as a non-recurrent Boltzmann machine.  The learning
rule derived from the interpretation as a non-recurrent Boltzmann machine
clearly does not involve backpropagation through time but is a proper
learning rule for the equivalent RNN.  As a result, training a DyBM is
free from the ``vanishing gradient problem'' \cite{vanishing, PMB13}.

The learning rule for some of the existing
recurrent neural networks involves STDP but in a more limited form than
our learning rule.  For example, the
learning rule of \cite{RNNwithSTDP} depends on the timing of spikes
but only on whether a post-synaptic neuron fires immediately after a
pre-synaptic neuron fires.  In our learning rule,
the magnitude of the changes in the weight can depend on the
difference between the timings of the two spikes, as has been observed
for biological neural networks \cite{abbott2000synaptic}.

An extended version of this paper has appeared in \cite{srep2015}.  This
paper, however, contains some of the details and perspectives that are
omitted from \cite{srep2015}.  Also, note that the notations and
terminologies in this paper are not necessarily consistent with those in
\cite{srep2015}.








\section{Defining dynamic Boltzmann machine}
\label{sec:DBM}

After reviewing the Boltzmann machine and the restricted Boltzmann
machine in Section~\ref{sec:DBM:BM}, we define the DyBM in
Section~\ref{sec:DBM:DBM}.  A learning rule for the DyBM and its
interpretation as STDP will be provided in Section~\ref{sec:STDP}.

\subsection{Conventional Boltzmann machine}
\label{sec:DBM:BM}

\begin{figure*}[t]
\begin{minipage}{0.33\linewidth}
\centering
\includegraphics[width=.66\linewidth]{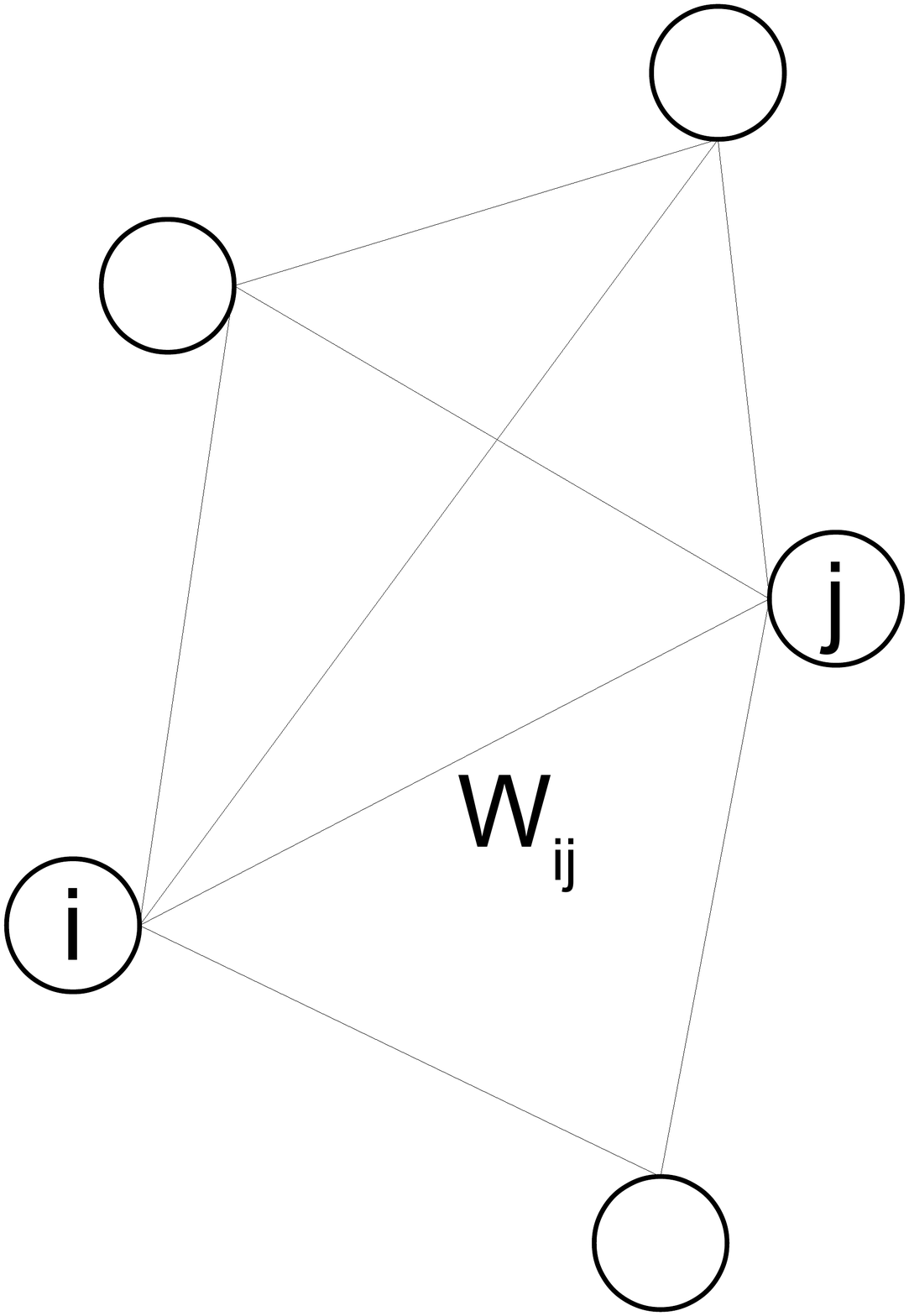}\\
(a) Boltzmann machine
\end{minipage}
\begin{minipage}{0.33\linewidth}
\centering
\includegraphics[width=.66\linewidth]{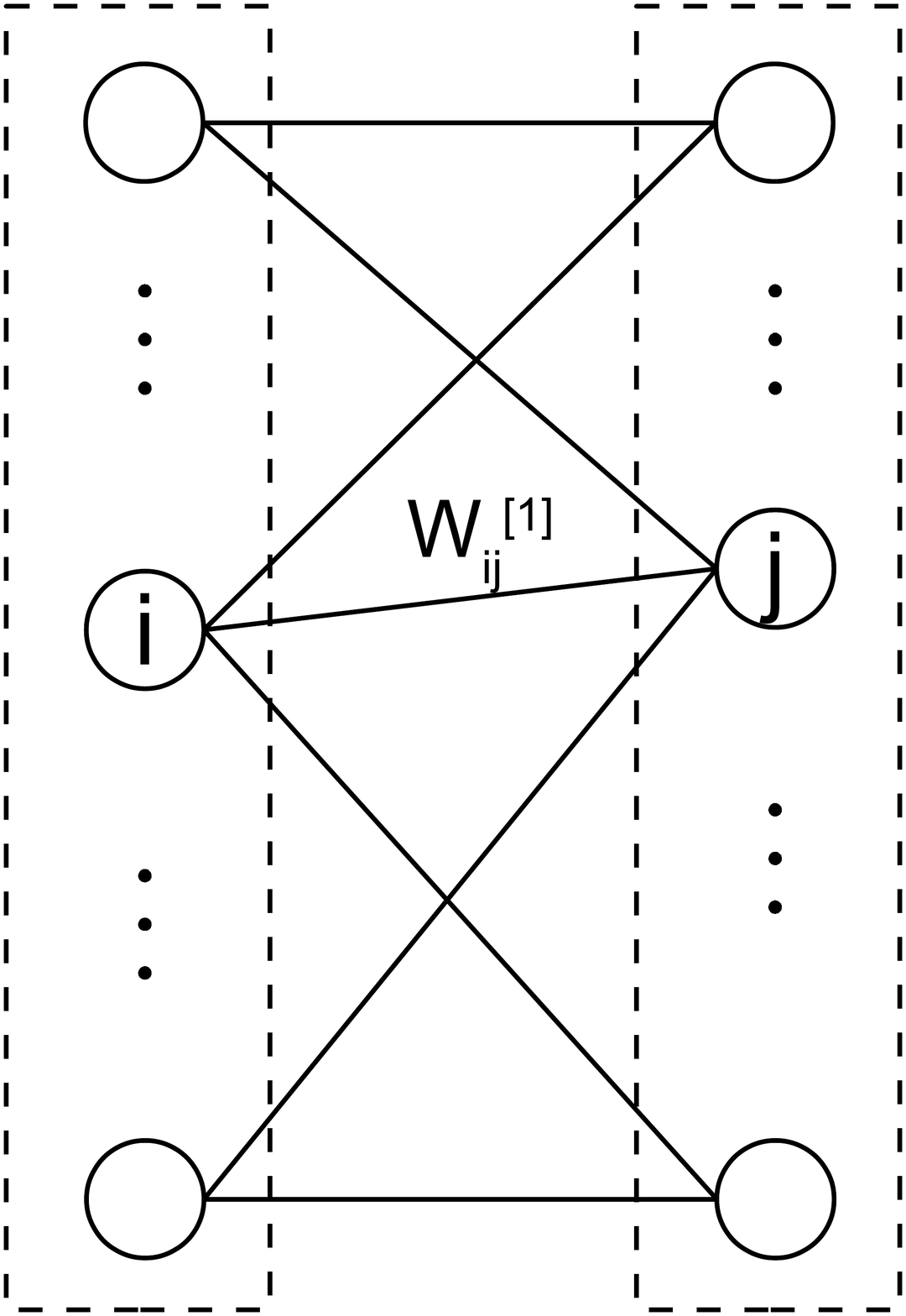}\\
(b) RBM
\end{minipage}
\begin{minipage}{0.33\linewidth}
\centering
\includegraphics[width=.66\linewidth]{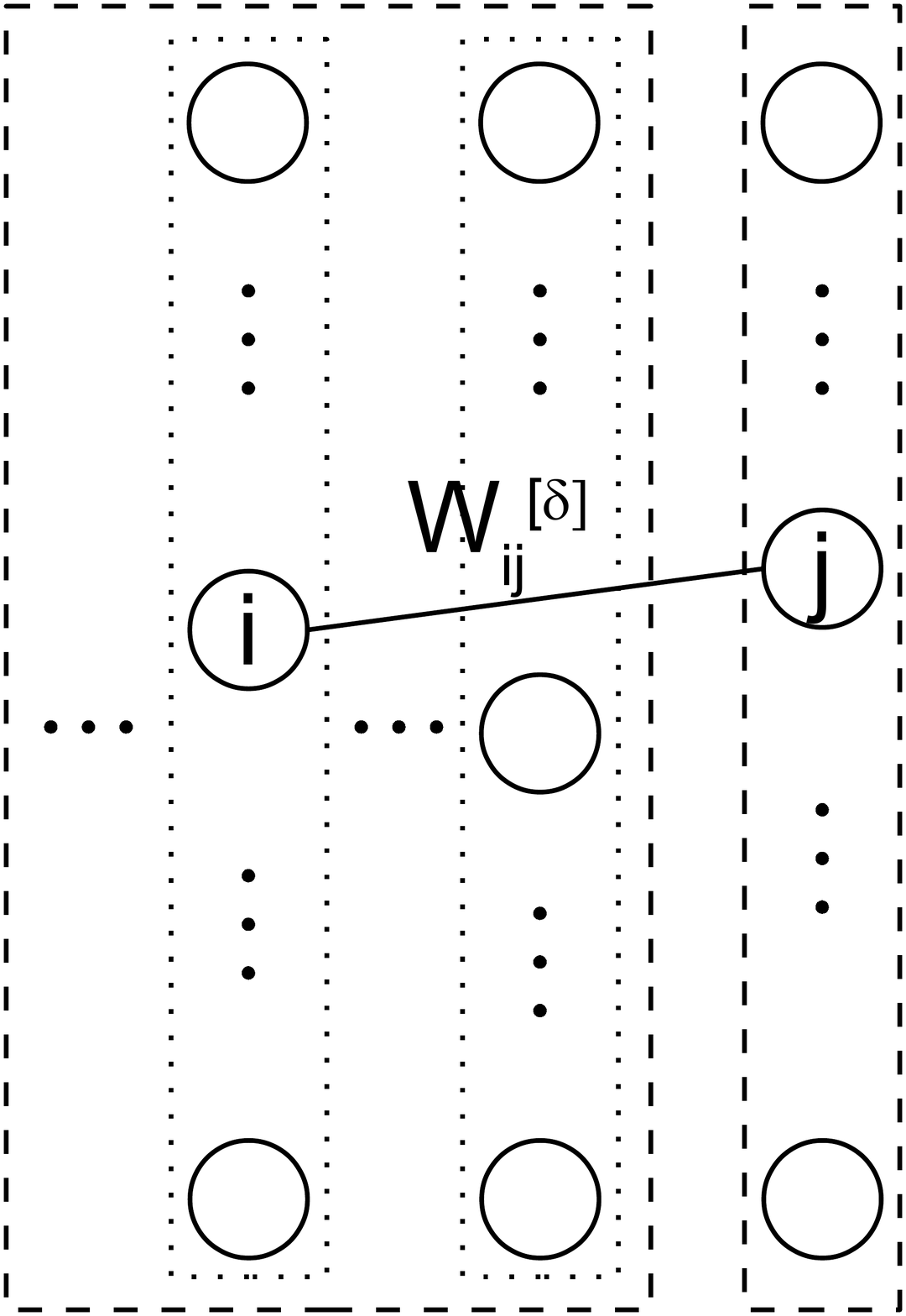}\\
(c) DyBM
\end{minipage}
\caption{(a) A Boltzmann machine, (b) a restricted Boltzmann machine, and (c) a DyBM.}
\label{fig:BM}
\end{figure*}

A Boltzmann machine is a network of units that are
mutually connected with weight (see Figure~\ref{fig:BM}~(a)).  Let $N$ be the number of units.
For $i\in[1,N]$, let $x_i$ be the value of the $i$-th unit, where
$x_i\in\{0,1\}$.  
Let $w_{i,j}$ be the weight between the $i$-th unit
and the $j$-th unit for $i,j\in[1,N]$.  It is standard to assume
that $w_{i,i}=0$ and $w_{i,j}=w_{j,i}$ for $i,j\in[1,N]$.  For
$i\in[1,N]$, the bias, $b_i$, is associated with the $i$-th unit.
We use the following notations of column vectors and matrices:
$\mathbf{x}\equiv(x_i)_{i\in[1,N]}$,
$\mathbf{b}\equiv(b_i)_{i\in[1,N]}$, and
$\mathbf{W}\equiv(w_{i,j})_{i,j\in[1,N]^2}$.  Let
$\theta\equiv(\mathbf{b},\mathbf{W})$ denote the parameters of the
Boltzmann machine.

The energy of the values, $\mathbf{x}$, for the $N$ units of
the Boltzmann machine having $\theta$ is given by
\begin{align}
E_\theta(\mathbf{x}) = - \mathbf{b}^\top \, \mathbf{x} - \frac{1}{2} \, \mathbf{x}^\top \, \mathbf{W} \, \mathbf{x}.
\label{eq:BMenergy}
\end{align}
The (equilibrium) probability of generating particular values, $\mathbf{x}$,
is given by
\begin{align}
P_\theta(\mathbf{x}) 
= \frac{\exp\left(-\tau^{-1} \, E_\theta(\mathbf{x})\right)}{\sum_{\mathbf{\tilde x}}\exp\left(-\tau^{-1} \, E_\theta(\mathbf{\tilde x})\right)},
\label{eq:BMprob}
\end{align}
where the summation over $\mathbf{\tilde x}$ denotes
the summation over all of the possible configurations of a binary vector of length $N$,
and $\tau$ is the parameter called temperature.

The Boltzmann machine can be trained in the direction of maximizing
the log likelihood of a given set, ${\cal D}$, of the vectors of the values:
\begin{align}
\nabla_\theta \log P_\theta({\cal D})
& = \sum_{\mathbf{x}\in{\cal D}} \nabla_\theta \log P_\theta(\mathbf{x}),
\end{align}
where $P_\theta({\cal D})\equiv\prod_{\mathbf{x}\in{\cal D}}P_\theta(\mathbf{x})$, and
\begin{align}
\nabla_\theta \log P_\theta(\mathbf{x})
& = - \tau^{-1} \Big( \! \nabla_\theta E_\theta(\mathbf{x}) -
 \sum_{\mathbf{\tilde x}} P_\theta(\mathbf{\tilde x}) \, \nabla_\theta
 E_\theta(\mathbf{\tilde x}) \! \Big).
\label{eq:BMlearn}
\end{align}
A Hebbian rule can be derived from (\ref{eq:BMlearn}).  For example,
to increase the likelihood of a data point, $\mathbf{x}$, we should update
$W_{i,j}$ by 
\begin{align}
W_{i,j} \leftarrow W_{i,j} + \eta \, (x_i \, x_j - \langle X_i \, X_j \rangle_\theta),
\label{eq:Hebb}
\end{align}
where $\langle X_i \, X_j \rangle_\theta$ denotes the expected value of the product of the values of the $i$-th unit and the $j$-th unit
with respect to $P_\theta$ in (\ref{eq:BMprob}), and
$\eta$ is a learning rate \cite{hinton1983optimal}.

A particularly interesting case is a restricted Boltzmann machine
(RBM), where the units are divided into two layers, and there is no
weight between the units in each layer (see Figure~\ref{fig:BM}~(b)).
In this case, $P_\theta(\mathbf{x})$ can be evaluated approximately
with contrastive divergence \cite{hinton2002training, SMH07}
or other methods \cite{Tie08, CRI11}
when the exact computation of
(\ref{eq:BMlearn}) is intractable (e.g., when $N$ is large).

A key property of the RBM that allows contrastive divergence is the
following conditional independence.  For $t\in[1,2]$, let
$\mathbf{x}^{[t]}$ be the values of the units in the $t$-th layer and
$\mathbf{b}^{[t]}$ be the corresponding bias.  Let $\mathbf{W}^{[1]}$
be the matrix whose $(i,j)$ element is the weight between the $i$-th
unit in the first layer and the $j$-th unit in the second layer.
Then the conditional probability of $\mathbf{x}^{[2]}$ given
$\mathbf{x}^{[1]}$ is given by
\begin{align}
 P_\theta(\mathbf{x}^{[2]}|\mathbf{x}^{[1]})
 &= \prod_{j\in[1,N]} P_{\theta,j}(x_j^{[2]}|\mathbf{x}^{[1]})\\
 &\equiv \prod_{j\in[1,N]} \frac{\exp\left(-\tau^{-1} \, E_{\theta,j}(x_j^{[2]}|\mathbf{x}^{[1]})\right)}{\displaystyle\sum_{x_j^{[2]}\in\{0,1\}}\exp\left(-\tau^{-1} \, E_{\theta,j}(x_j^{[2]}|\mathbf{x}^{[1]})\right)},
\label{eq:conditional_independence}
\end{align}
where $P_{\theta,j}(x_j^{[2]}|\mathbf{x}^{[1]})$ denotes the conditional probability that the $j$-th unit of
the second layer has value $x_j^{[2]}$ given that the first layer has values $\mathbf{x}^{[1]}$, and
we define
\begin{align}
E_{\theta,j}(x_j^{[2]}|\mathbf{x}^{[1]})
\equiv - b_j \,x_j^{[2]} - (\mathbf{x}^{[1]})^\top \, W_{:,j}^{[1]} \, x_j^{[2]},
\end{align}
where $W_{:,j}^{[1]}$ denotes the $j$-th column of $\mathbf{W}^{[1]}$.
Namely, the value of the units in the second layer, $\mathbf{x}^{[2]}$, is conditionally independent of each other given $\mathbf{x}^{[1]}$.

\subsection{Dynamic Boltzmann machine}
\label{sec:DBM:DBM}

We propose the dynamic Boltzmann machine
(DyBM), which can have infinitely many layers of units (see
Figure~\ref{fig:BM}~(c)).  Similar to the RBM, the DyBM has no weight
between the units in the right-most layer of Figure~\ref{fig:BM}~(c).
Unlike the RBM, each layer of the DyBM has a common number, $N$, of
units, and the bias and the weight in the DyBM can be shared among
different units in a particular manner.  

Formally, we define the DyBM-$T$ as the Boltzmann machine having $T$
layers from $-T+1$ to $0$, where $T$ is a positive integer or
infinity.  Let $\mathbf{x}\equiv(\mathbf{x}^{[t]})_{-T<t\le 0}$, where
$\mathbf{x}^{[t]}$ is the values of the units in the $t$-th layer,
which we consider as the values at time $t$.  Let $\mathbf{b}$ be the
bias for the $N$ units at the 0-th layer (the right-most layer of
Figure~\ref{fig:BM}~(c)).  For $\delta\ge 1$, let
$\mathbf{W}^{[\delta]}$ be the matrix whose $(i,j)$ element,
$W_{i,j}^{[\delta]}$, denotes the weight between the $i$-th unit at
time $-\delta$ and the $j$-th unit at time $0$ for any $t$.  A unit in
the $s$-th layer for $s\le -1$ can have arbitrary bias and arbitrary
weight with a unit in the $t$-th layer for $t\le -1$,
but such bias and weight
have no effect in the DyBM, which we will see in the following.  The
DyBM with infinitely many layers is defined with a formal limit of
$T\to\infty$.

For a DyBM-$T$, consider $P_\theta(\mathbf{x}^{[0]} |
\mathbf{x}^{(-T,-1]})$, the conditional probability of
  $\mathbf{x}^{(0)}$ given $\mathbf{x}^{(-T,-1]}$, where we use
    $\mathbf{x}^{I}$ for an interval $I$ such as $(-T,-1]$ to denote
      $(\mathbf{x}^{[t]})_{t\in I}$.  Because the units in the 0-th layer
have no weight with each other, this conditional probability has
      the property of conditional independence analogous to
      (\ref{eq:conditional_independence}).  Note that
      $P_\theta(\mathbf{x}^{[0]} | \mathbf{x}^{(-T,-1]})$ is
      independent of the bias and the weight
that are not associated with the units in the 0-th layer.



We propose the DyBM as a model of a time-series in the following sense.
Specifically, given a history $\mathbf{x}^{(-T,-1]}$ of a time-series,
the DyBM-$T$ gives the probability of the next values,
$\mathbf{x}^{[0]}$ of the time-series with $P_\theta(\mathbf{x}^{[0]}
| \mathbf{x}^{(-T,-1]})$.  A DyBM-2 can be interpreted as a Markov
model, and DyBM-$T$ for $T>2$ as a $(T-1)$-st order Markov model, where
the next values are conditionally independent of the history given the
values of the last $(T-1)$ steps.  With a DyBM-$\infty$, the next
values can depend on the whole history of the time-series.  In
principle, the DyBM-$\infty$ can thus model any time-series
possibly
with long-term dependency, as long as the values of the time-series
at a moment is conditionally independent of each other given its values
preceding that moment.  Using the conditional probability
given by a DyBM-$T$, the probability of a sequence,
$\mathbf{x}=\mathbf{x}^{(-L,0]}$, of length $L$ is given by
\begin{align}
p(\mathbf{x}) = \prod_{t=-L+1}^0 P_\theta(\mathbf{x}^{[t]}| \mathbf{x}^{(t-T,:t-1]}),
\label{eq:chain}
\end{align}
where we arbitrarily define $\mathbf{x}^{[t]}\equiv\mathbf{0}$ for $t\le -L$.
Namely, the values are set zero if there are no corresponding history.


\section{Spike-timing dependent plasticity}
\label{sec:STDP}

We derive a learning rule for a DyBM-$T$ in such a way that the log
likelihood of a given (set of) time-series is maximized.  We will see
that the learning rule is particularly simplified in the limit of
$T\to\infty$ when the parameters of the DyBM-$\infty$ have particular
structures.  We will show that this learning rule exhibits various
characteristics of spike-timing dependent plasticity (STDP).

\subsection{General learning rule of dynamic Boltzmann machines}
\label{sec:STDP:general}

The log likelihood of a given set, ${\cal D}$, of time-series can be
maximized by maximizing the sum of the log likelihood of
$\mathbf{x}\in{\cal D}$.  By (\ref{eq:chain}), the log likelihood of
$\mathbf{x}=\mathbf{x}^{(-L,0]}$ has the following
gradient:
\begin{align}
\nabla_\theta \log p(\mathbf{x}) = \sum_{t=-L+1}^0 \nabla_\theta \log P_\theta(\mathbf{x}^{[t]}| \mathbf{x}^{(t-T,t-1]}).
\label{eq:grad}
\end{align}
We can thus exploit the conditional independence of (\ref{eq:conditional_independence})
to derive the learning rule:
\begin{align}
\lefteqn{\hspace{-5mm}\nabla_\theta \log P_\theta(\mathbf{x}^{[0]}|
 \mathbf{x}^{(-T,-1]})} \notag\\
= &-\tau^{-1} \, \sum_{j\in[1,N]} \bigg( \nabla_\theta E_{\theta,j}(x_j^{[0]}|\mathbf{x}^{(-T,-1]}) 
 - \!\!\!\! \sum_{\tilde x_j^{[0]}\in\{0,1\}} \!\!\!\! P_{\theta,j}(\tilde x_j^{[0]}|\mathbf{x}^{(-T,-1]}) \, \nabla_\theta E_{\theta,j}(\tilde x_j^{[0]}|\mathbf{x}^{(-T,-1]})\bigg).
\label{eq:grad0}
\end{align}
One could, for example, update $\theta$ in the direction of
(\ref{eq:grad0}) every time new $\mathbf{x}^{[0]}$ is observed, using
the latest history, $\mathbf{x}^{(-T,-1]}$.  This is an approach of
  stochastic gradient.  In practice, however, the computation of
  (\ref{eq:grad0}) can be intractable for a large $T$, because there
  are $\Theta(M\,T)$ parameters to learn, where $M$ is the number of
  the pairs of connected units ($M=\Theta(N^2)$ when all of the units
  are densely connected).

\subsection{Deriving a specific learning rule}
\label{sec:STDP:specific}

We thus propose a particular form of weight sharing, which is
motivated by observations from biological neural networks
\cite{abbott2000synaptic} but leads to particularly simple, exact, and
efficient learning rule.  In biological neural networks, STDP has been
postulated and supported experimentally.  In particular, the synaptic
weight from a pre-synaptic neuron to a post-synaptic neuron is
strengthened, if the post-synaptic neuron fires (generates a spike)
shortly {\em after} the pre-synaptic neuron fires (i.e., long term
potentiation or LTP).  This weight is weakened, if the post-synaptic
neuron fires shortly {\em before} the pre-synaptic neuron fires (i.e.,
long term depression or LTD).  These dependency on the timing of
spikes is missing in the Hebbian rule for the Boltzmann machine
(\ref{eq:Hebb}).

\begin{figure}[t]
\centering \includegraphics[width=0.5\linewidth]{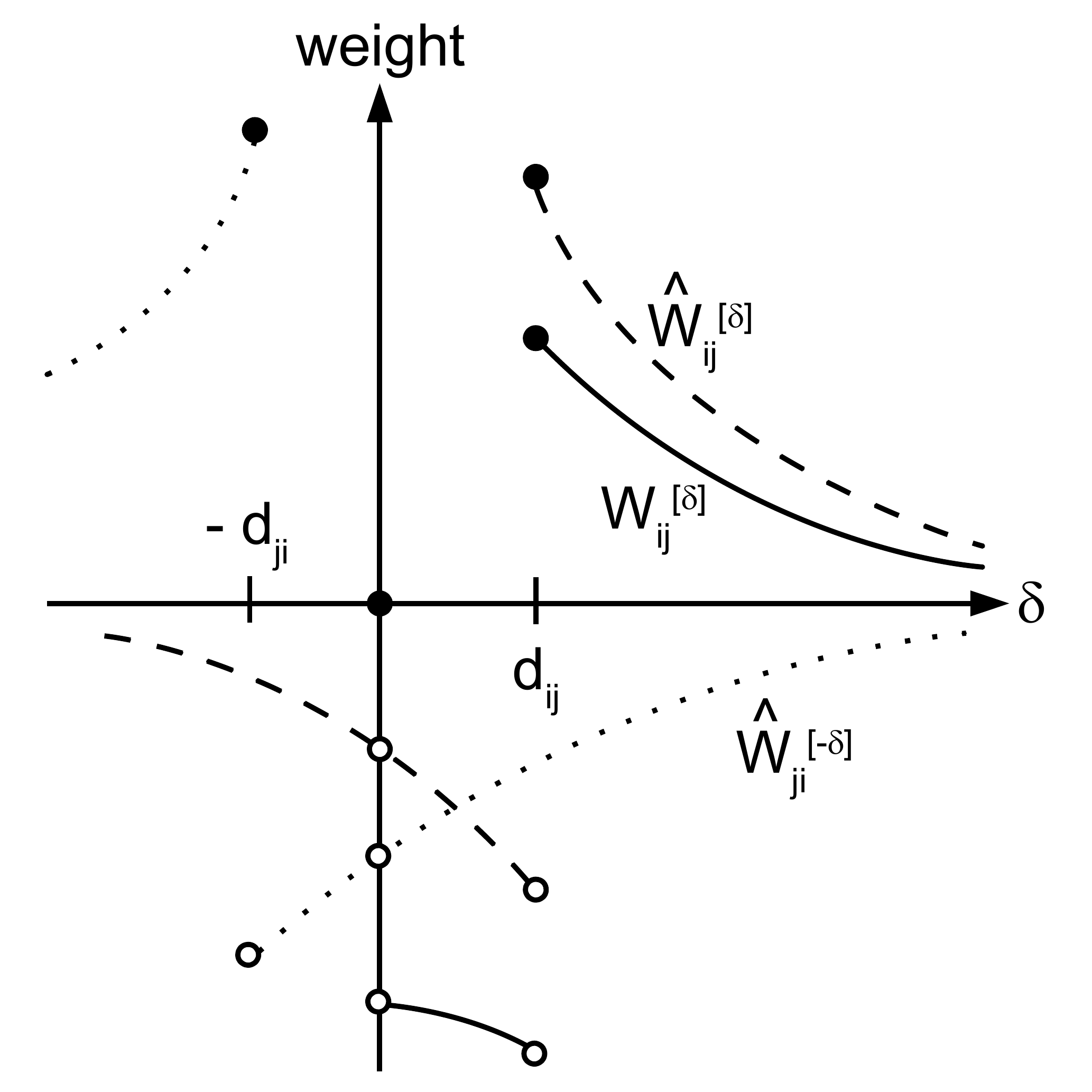}
 \caption{
The figure illustrates Equation~(\ref{eq:W}) with particular forms of
Equation~(\ref{eq:geometric}).  The horizontal axis represents $\delta$,
and the vertical axis represents the value of $W_{i,j}^{[\delta]}$ (solid
curves), $\hat W_{i,j}^{[\delta]}$ (dashed curves), or $\hat
W_{j,i}^{[-\delta]}$ (dotted curves).  Notice that $W_{i,j}^{[\delta]}$ is
defined for $\delta > 0$ and is discontinuous at $\delta = d_{i,j}$.  On the
other hand, $\hat W_{i,j}^{[\delta]}$ and $\hat W_{j,i}^{[-\delta]}$ are
defined for $-\infty < \delta < \infty$ and discontinuous at $\delta =
d_{i,j}$ and $\delta = -d_{j,i}$, respectively.}
\label{fig:weight}
\end{figure}

To derive a learning rule that has the characteristics of STDP with LTP and LTD, we
consider the weight of the form illustrated in
Figure~\ref{fig:weight}.  For $\delta>0$, we define the weight,
$W_{i,j}^{[\delta]}$, as the sum of two weights, $\hat
W_{i,j}^{[\delta]}$ and $W_{j,i}^{[-\delta]}$:
\begin{align}
W_{i,j}^{[\delta]} = \hat W_{i,j}^{[\delta]} + \hat W_{j,i}^{[-\delta]}. 
\label{eq:W}
\end{align}

In Figure~\ref{fig:weight},
the value of $\hat W_{i,j}^{[\delta]}$ is high when $\delta=d_{i,j}$,
the (synaptic) delay from $i$-th (pre-synaptic) unit to the $j$-th
(post-synaptic) unit.  Namely, the
post-synaptic neuron is likely to fire (i.e., $x_j^{[0]}=1$)
immediately after the spike from the pre-synaptic unit arrives with
the delay of $d_{i,j}$ (i.e, $x_i^{[-d_{i,j}]}=1$).  This likelihood is
controlled by the magnitude of $\hat W_{i,j}^{[d_{i,j}]}$, which we will
learn from training data.  The value of $\hat W_{i,j}^{[\delta]}$ gradually
decreases, as $\delta$ increases from $d_{i,j}$.  That is, the effect
of the stimulus of the spike arrived from the $i$-th unit diminishes
with time \cite{abbott2000synaptic}.

The value of $\hat W_{i,j}^{[d_{i,j}-1]}$ is low, suggesting that the
post-synaptic unit is unlikely to fire (i.e., $x_j^{[0]}=1$)
immediately {\em before} the spike from the $i$-th (pre-synaptic) unit
arrives.  This unlikelihood is controlled by the magnitude of $\hat
W_{i,j}^{[d_{i,j}-1]}$, which we will learn.  As $\delta$ decreases
from $d_{i,j}-1$, the magnitude of $\hat W_{i,j}^{[\delta]}$ gradually
decreases \cite{abbott2000synaptic}.  Here, $\delta$ can get smaller
than 0, and $\hat W_{i,j}^{[\delta]}$ with $\delta<0$ represents the
weight between the spike of the pre-synaptic neuron that is generated
after the spike of the post-synaptic neuron.

The assumption of $W_{i,j}^{[0]} = 0$ is
convenient for computational purposes but can be justified in the
limit of infinitesimal time steps.  Specifically, consider a scaled
DyBM where both the step size of the time and the probability of
firing are made $1/n$-th of the original DyBM.  In the
limit of $n\to\infty$, the scaled DyBM has continuous time, and the
probability of having simultaneous spikes from two units tends to
zero.

\begin{figure}[t]
 \centering
\includegraphics[width=.5\linewidth]{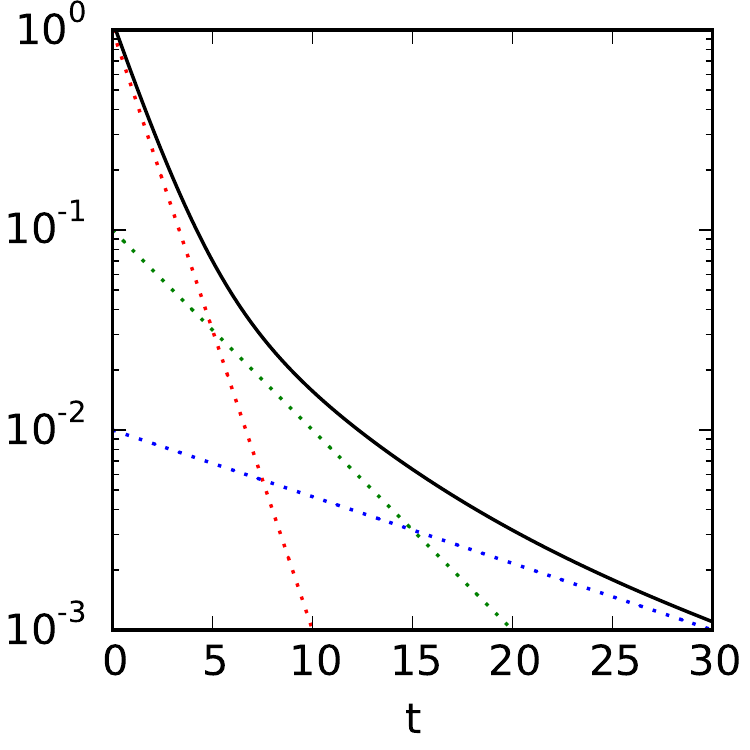}
\caption{The solid curve shows the sum of three geometric functions
 ($10^{-0.3t}$, $10^{-0.1t-1}$, and $10^{-0.1t/3-2}$)
 shown with dotted lines (the vertical axis is in the log scale).}
\label{fig:exp}
\end{figure}

For tractable learning and inference, we assume the following form of weight:
\begin{align}
\hat W_{i,j}^{[\delta]} = \left\{\begin{array}{ll}
0 & \mbox{if } \delta = 0\\
\displaystyle\sum_{k\in{\cal K}} u_{i,j,k} \, \lambda_k^{\delta-d_{i,j}} & \mbox{if } \delta \ge d_{i,j}\\
\displaystyle\sum_{\ell\in{\cal L}} -v_{i,j,\ell} \, \mu_\ell^{-\delta} & \mbox{otherwise}
\end{array}\right.
\label{eq:geometric}
\end{align}
where $\lambda_k,\mu_\ell\in(0,1)$ for $k\in{\cal K}$ and $\ell\in{\cal
L}$.  We will learn the values of $u_{i,j,k}$ and $v_{i,j,\ell}$ based
on training dataset.  We assume that $\lambda_k$ for $k\in{\cal K}$,
$\mu_\ell$ for $\ell\in{\cal L}$, and $d_{i,j}$ for $i,j\in[1,N]$ are
given (or need to be learned as hyper-parameters).  With an analogy to
biological neural networks, these given parameters ($\lambda_k$,
$\mu_\ell$, and $d_{i,j}$) are determined based on physical constraints
or chemical properties, while the weight ($u_{i,j,k}$ and
$v_{i,j,\ell}$) and the bias ($\mathbf{b}$) are learned based on the
neural spikes ($\mathbf{x}$).  The sum of geometric functions with
varying decay rates \cite{Killeen01} in (\ref{eq:geometric}) is
motivated by long-term memory (or dependency)
\cite{MW13,sutskever2008recurrent}.  See Figure~\ref{fig:exp} for the
flexibility of the sum of geometric functions.  In particular, the sum
of geometric functions can well approximate a hyperbolic function, whose
value decays more slowly than any geometric functions.  This slow decay
is considered to be essential for long-term memory.  However, our
results also hold for the simple cases where $|{\cal K}|=|{\cal L}|=1$.

With the above specifications and letting $T\to\infty$, we can now represent the
$E_{\theta,j}(x_j^{[0]}|\mathbf{x}^{(-T,-1]})$ appearing in (\ref{eq:grad0}) as follows:
\begin{align}
E_{\theta,j}(x_j^{[0]}|\mathbf{x}^{(-\infty,-1]})
& = - b_j \, x_j^{[0]} - \sum_{t=-\infty}^{-1} (\mathbf{x}^{[t]})^\top \, W_{:,j}^{[-t]} \, x_j^{[0]},
\label{eq:Ej}
\end{align}
where $W_{:,j}^{[-t]} \equiv \big(W_{i,j}^{[-t]}\big)_{i=1,\ldots,N}$ denotes a column vector.
By (\ref{eq:W}) and (\ref{eq:geometric}), the second term of (\ref{eq:Ej}) is given by
\begin{align}
 \sum_{t=-\infty}^{-1} (\mathbf{x}^{[t]})^\top \, W_{:,j}^{[-t]}
 \, x_j^{[0]}
& = \sum_{i=1}^N \! \Big( \! 
\sum_{k\in{\cal K}} \! u_{i,j,k}  \alpha_{i,j,k}
\! - \! \sum_{\ell\in{\cal L}} \! v_{i,j,\ell}  \beta_{i,j,\ell}
\! - \! \sum_{\ell\in{\cal L}} \! v_{j,i,\ell} , \gamma_{i,\ell}
\Big) x_j^{[0]}
\end{align}
where $\alpha_{i,j,k}$, $\beta_{i,j,\ell}$, and $\gamma_{i,\ell}$ are
the quantities, which we refer to as eligibility traces, that depend on
$x_i^{(-\infty,-1]}$, the history of the $i$-th unit:
\begin{align}
\alpha_{i,j,k} & \equiv \sum_{t=-\infty}^{-d_{i,j}} \lambda_k^{-t-d_{i,j}} \, x_i^{[t]}; 
\label{eq:eligibility1} \\
\beta_{i,j,\ell} & \equiv \sum_{t=-d_{i,j}+1}^{-1} \mu_\ell^{t} \, x_i^{[t]}; \\
\gamma_{i,\ell} & \equiv \sum_{t=-\infty}^{-1} \mu_\ell^{-t} \, x_i^{[t]}.\
\label{eq:eligibility3}
\end{align}

We now derive the derivatives of (\ref{eq:Ej}), which we need for (\ref{eq:grad0}), as follows:\\
\begin{align}
\frac{\partial}{\partial b_j} E_{\theta,j}(x_j^{[0]}|\mathbf{x}^{(-\infty,-1]})
& = - x_j^{[0]}, \\
\frac{\partial}{\partial u_{i,j,k}} E_{\theta,j}(x_j^{[0]}|\mathbf{x}^{(-\infty,-1]})
& = - \alpha_{i,j,k} \, x_j^{[0]}, \\
\frac{\partial}{\partial v_{i,j,\ell}} E_{\theta,j}(x_j^{[0]}|\mathbf{x}^{(-\infty,-1]})
& = \beta_{i,j,\ell} \, x_j^{[0]}, \\
\frac{\partial}{\partial v_{j,i,\ell}} E_{\theta,j}(x_j^{[0]}|\mathbf{x}^{(-\infty,-1]})
& = \gamma_{i,\ell} \, x_j^{[0]}.
\end{align}

By plugging the above derivatives into (\ref{eq:grad0}), we have, for
$i,j\in[1,N]$, $k\in{\cal K}$, and $\ell\in{\cal L}$, that
\begin{align}
\frac{\partial}{\partial b_j} \log P_{\theta}(\mathbf{x}^{[0]}|\mathbf{x}^{(-\infty,-1]})
 & = \tau^{-1} \, \big(x_j^{[0]} - \langle X_j^{[0]}\rangle_\theta\big)
\label{eq:learn_b}\\
 \frac{\partial}{\partial u_{i,j,k}} \log
 P_{\theta}(\mathbf{x}^{[0]}|\mathbf{x}^{(-\infty,-1]})
& = \tau^{-1} \, \alpha_{i,j,k} \, \big(x_j^{[0]} - \langle X_j^{[0]}\rangle_\theta\big) 
\label{eq:learn_u}\\
 \frac{\partial}{\partial v_{i,j,\ell}} \log
 P_{\theta}(\mathbf{x}^{[0]}|\mathbf{x}^{(-\infty,-1]})
& = -\tau^{-1} \, \beta_{i,j,\ell} \, \big(x_j^{[0]} - \langle X_j^{[0]}\rangle_\theta\big)
-\tau^{-1} \, \gamma_{j,\ell} \, \big(x_i^{[0]}- \langle X_i^{[0]}\rangle_\theta\big),
\label{eq:learn_v}
\end{align}
where $\langle X_j^{[0]}\rangle_\theta$ denotes the expected value of the $j$-th unit in the 0-th layer
of the DyBM-$\infty$ 
given the history $\mathbf{x}^{(-\infty,-1]}$.  Because the value is binary, this expected value is
given by 
\begin{align}
\langle X_j^{[0]}\rangle_\theta
&= P_{\theta,j}(1|\mathbf{x}^{(-\infty,-1]})\\
&= \frac{\exp\left(-\tau^{-1} \, E_{\theta,j}(1|\mathbf{x}^{(-\infty,-1]})\right)}{1+\exp\left(-\tau^{-1} \, E_{\theta,j}(1|\mathbf{x}^{(-\infty,-1]})\right)},
\label{eq:EXj}
\end{align}
which can be calculated, using the eligibility traces in
(\ref{eq:eligibility1})-(\ref{eq:eligibility3}).
Here, the first term of the denominator of (\ref{eq:EXj}) is 1, because
$E_{\theta,j}(0|\mathbf{x}^{(-\infty,-1]})=0$.
To maximize the likelihood of a given set, ${\cal D}$, of sequences, the parameters $\theta$ can be updated with
\begin{align}
\theta 
\leftarrow \theta + \eta \, \sum_{\mathbf{x}\in{\cal D}} \nabla_\theta \log P_\theta(\mathbf{x}^{[0]}|\mathbf{x}^{(-\infty,-1]}).
\end{align}

Typically, a single time-series, $\mathbf{y}^{[1,L]}$, is available
for training the DyBM-$\infty$.  In this case, we form
${\cal D}\equiv\{ \mathbf{y}^{[1,t]} \mid t\in[1,L] \}$,
where $\mathbf{y}^{[1,t]}$ is used as
$\mathbf{x}^{[0]}\equiv\mathbf{y}^{[t]}$ and
$\mathbf{x}^{(-\infty,-1]}\equiv\mathbf{y}^{(-\infty,t-1]}$, where
    recall that we arbitrarily set zeros when there are no history (i.e.,
    $\mathbf{y}^{[t]}\equiv\mathbf{0}$ for $t\le 0$).

When ${\cal D}$ is made from a single time-series, the eligibility traces (\ref{eq:eligibility1})-(\ref{eq:eligibility3})
needed for training with $\mathbf{y}^{[1,t]}$
can be computed recursively from the ones used for $\mathbf{y}^{[1,t-1]}$.
In particular, we have
\begin{align}
\alpha_{i,j,k} &\leftarrow \lambda_k \, \left( \alpha_{i,j,k} + y_i^{[t-d_{i,j}]} \right) \\
\gamma_{i,\ell} &\leftarrow \mu_\ell \, \left( \gamma_{i,\ell} + y_i^{[t-1]}\right).
\end{align}
This recursive calculation requires keeping the following FIFO queue of length $d_{i,j}-1$:
\begin{align}
q_{i,j} \equiv \left(y_i^{[t-2]},\ldots,y_i^{[t-d_{i,j}+1]},y_i^{[t-d_{i,j}]}\right)
\end{align}
for each $i,j\in[1,N]$.  After training with $\mathbf{y}^{[1,t-1]}$, the
$q_{i,j}$ is updated by adding $y_i^{[t-1]}$ and deleting
$y_i^{[t-d_{i,j}]}$.  
The remaining
eligibility trace $\beta_{i,j,\ell}$ can be calculated
non-recursively by the use of $q_{i,j}$.
In fact, our experience suggests that a recursive calculation
of $\beta_{i,j,k}$ is amenable to numerical instability, so that
$\beta_{i,j,k}$ should be calculated non-recursively.


\subsection{Homogeneous dynamic Boltzmann machine}

\begin{figure}[t]
\centering \includegraphics[width=0.5\linewidth]{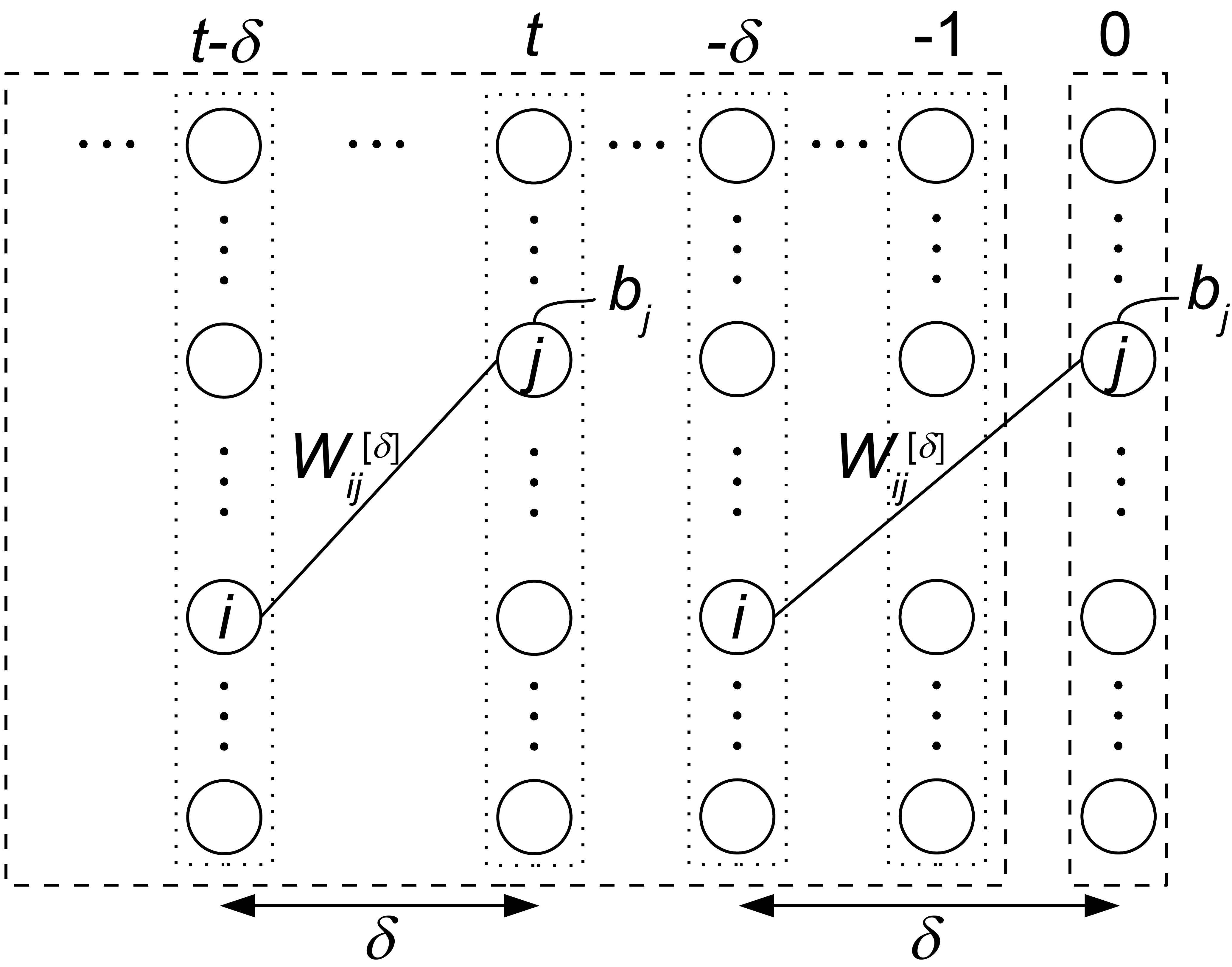}
 \caption{The homogeneous DyBM.}
\label{fig:homogeneous}
\end{figure}

Here, we will show that the DyBM-$\infty$ can indeed be understood as a
generative model of a time series.  For this purpose, we will specify the
bias for the units in the $s$-th layer for $s\le-1$ and the weight
between the units in the $s$-th layer and the units in the $t$-th layer,
for $s,t\le-1$.  Recall that these bias and weight have not been
specified in the discussion so far, because they do not affect the
conditional probability:
\begin{align}
P_\theta(\mathbf{x}^{[0]}|\mathbf{x}^{(-\infty,-1]}).
 \label{eq:conditional_probability}
\end{align}
This model of a single time step with (\ref{eq:conditional_probability})
is used iteratively in (\ref{eq:chain}) to define the distribution of a
time series.  Strictly speaking, however, this iterative use of
(\ref{eq:conditional_probability}) is not a generative model defined
solely with a Boltzmann machine.

We now consider the following homogeneous DyBM (see
Figure~\ref{fig:homogeneous}), which is a special case of a DyBM-$\infty$.  Let each layer of the units has a common
vector of bias.  That is, for any $s$, the units in the $s$-th
layer have the bias, $\mathbf{b}$.  Let the matrix of the weight between
two layers, $s$ and $t$, depend only on the distance, $t-s$, between the
two layers.  That is, for any pair of $s$ and $t$, the $i$-th unit in the $s$-th
layer and the $j$-th unit in the $t$-th layer is connected with the
weight, $W_{i,j}^{[t-s]}$, for $i, j\in[1,N]$.  The learning rule derived
for the DyBM-$\infty$
in Section~\ref{sec:STDP:general} - Section~\ref{sec:STDP:specific}
holds for the homogeneous DyBM.

The key property of the homogeneous DyBM is that the homogeneous DyBM
consisting of the layers up to the $t$-th layer, for $t<0$, is
equivalent to the homogeneous DyBM consisting of the layers up to the
$0$-th layer.  Therefore, the iterative use of the model of the single
time step (\ref{eq:conditional_probability}) is now equivalent to the
generative model of a time-series defined with a single homogeneous
DyBM.  Specifically, the values of the time-series at time $t$ are
generated based on the conditional probability,
$P_\theta(\mathbf{x}^{[t]}|\mathbf{x}^{(-\infty,t-1]})$, that is given
by the homogeneous DyBM consisting of the layers up to the $t$-th layer.
The values, $\mathbf{x}^{[t]}$, generated at time $t$ are then used as a
part of the history $\mathbf{x}^{(-\infty,t]}$ for the homogeneous DyBM
consisting of the layers up to the $(t+1)$-st layer, which in turn
defines the conditional probability of the values at time $t+1$.  The
homogeneous DyBM can also have the layers for positive time steps ($t>0$).
The homogeneous DyBM can then be interpreted as a recurrent neural network,
which we will discuss in the following with reference to an artificial
neural network.

\subsection{Interpretation as an artificial neural network}

\begin{figure}
\centering
\includegraphics[width=0.5\linewidth]{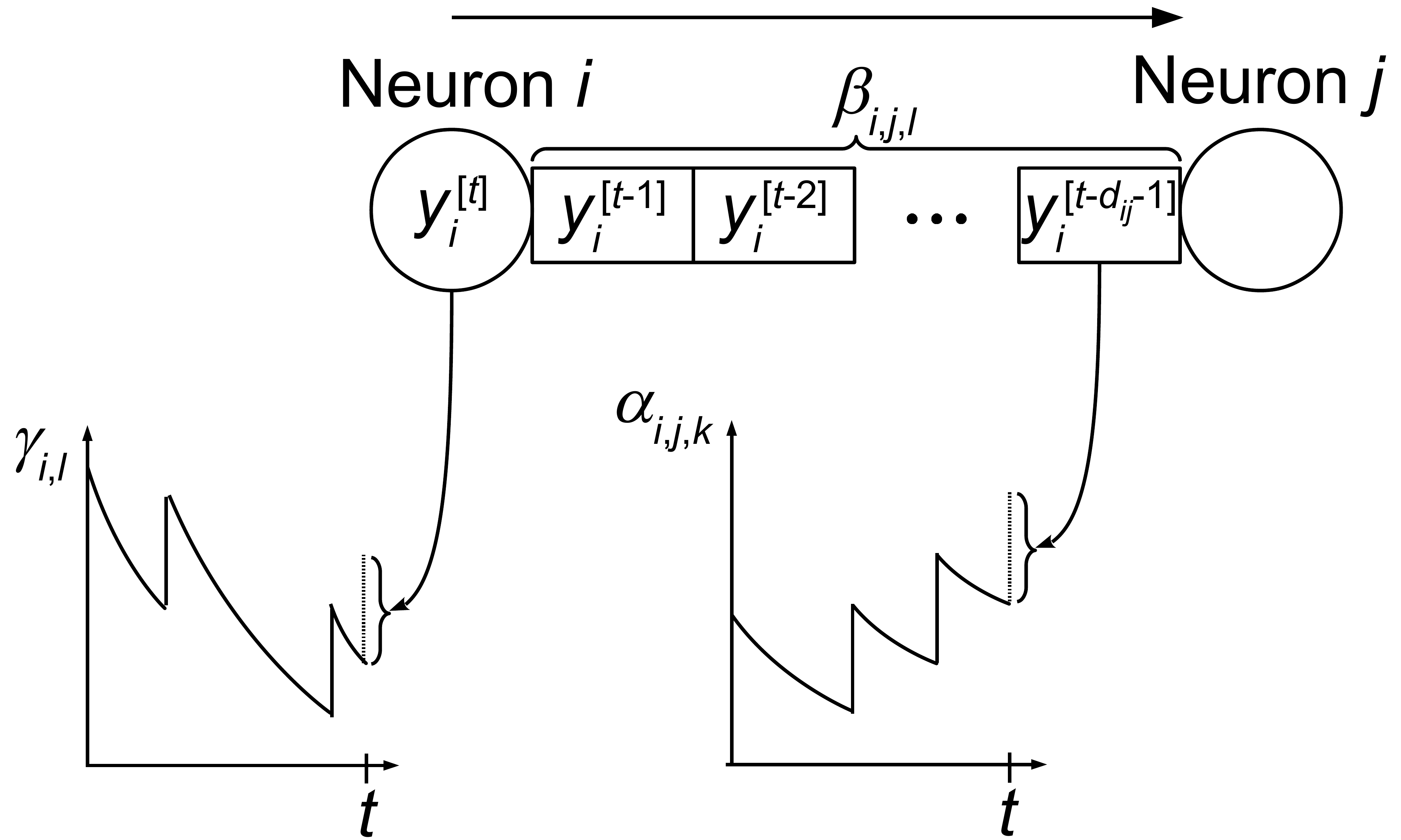}
\caption{Spikes traveling from a pre-synaptic neuron ($i$) to a post-synaptic neuron
($j$) and eligibility traces.}
\label{fig:neuron}
\end{figure}

Figure~\ref{fig:neuron} illustrates the learning rule derived in
Section~\ref{sec:STDP:general}-Section~\ref{sec:STDP:specific} from a
point of artificial neural networks.  Consider a pre-synaptic neuron,
$i$, and a post-synaptic neuron, $j$.  The FIFO queue, $q_{i,j}$, can
be considered as an axon that stores the spikes traveling from $i$
to $j$.  The conduction delay of this axon is $d_{i,j}$, and the
spikes generated in the last $d_{i,j}-1$ steps are stored.  The spikes in the
axon determine the value of 
$\beta_{i,j,\ell}$ for each $\ell\in{\cal L}$.  Another eligibility
trace, $\gamma_{i,\ell}$, records the aggregated information about the
spikes generated at the neuron $i$, where the spikes
generated in the past are discounted with the rate that depends on
$\ell\in{\cal L}$.  The remaining eligibility trace, $\alpha_{i,j,k}$,
records the aggregated information about the spikes that have reached
$j$ from $i$, where the spikes arrived in the past are discounted with
the rate that depends on $k\in{\cal K}$.

\begin{figure}
 \centering
\includegraphics[width=0.5\linewidth]{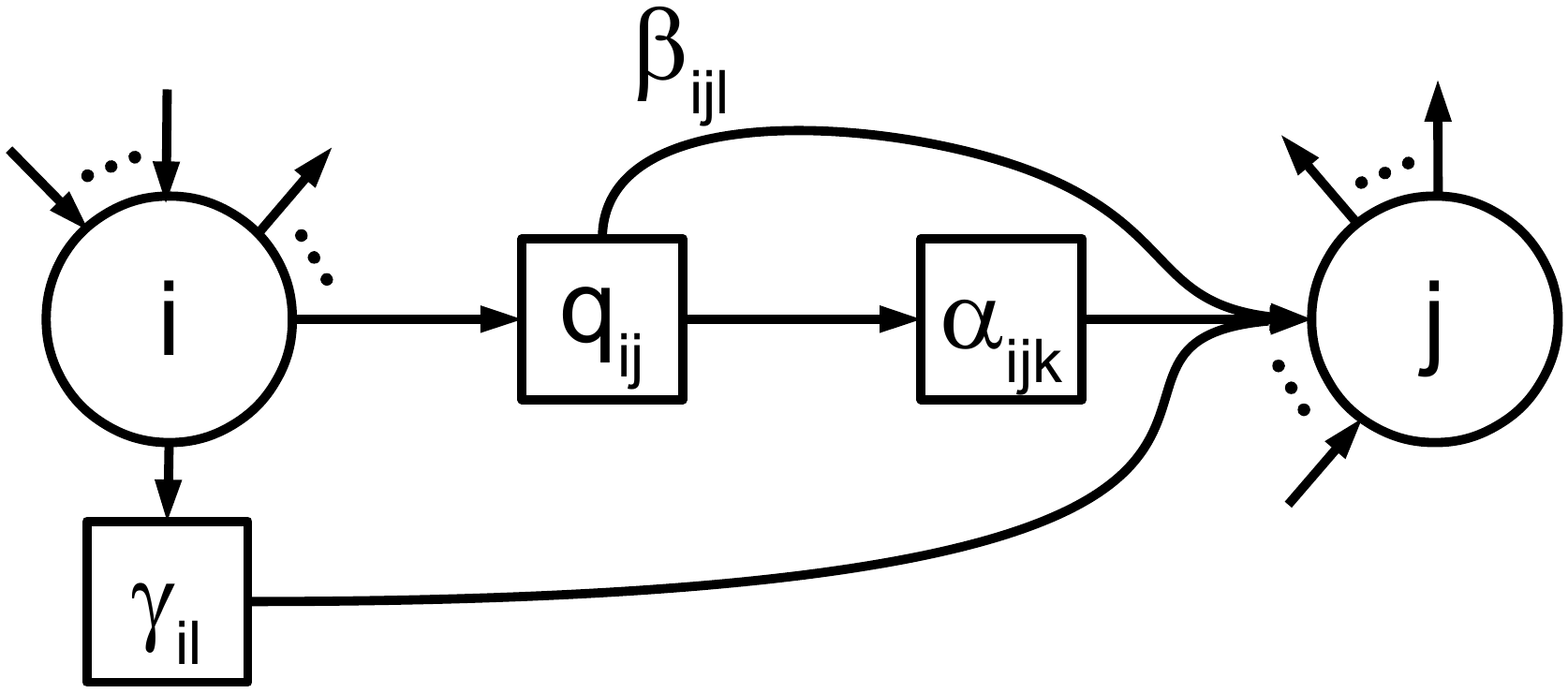}
\caption{The homogeneous DyBM as a recurrent neural network with memory.}
\label{fig:RNN}
\end{figure}

The DyBM-$\infty$ can then be considered as a recurrent neural network,
taking binary values,
equipped with memory units that store eligibility traces and the FIFO
queue (see Figure~\ref{fig:RNN}).  For learning or inference with an
$N$-dimensional binary time-series of arbitrary length, this
recurrent neural network needs the working space of $O(N+M\,D)$ binary
bits and $O(M\,|{\cal K}| + M\,|{\cal L}|)$ floating-point numbers,
where $M$ is the number of ordered pairs of connected units (i.e, the
number of the pairs of $i$ and $j$ such that $W_{i,j}^{[\delta]}\neq
0$ for a $\delta\ge 1$ in the DyBM-$\infty$), and $D$ is the maximum
delay such that $d_{i,j}\le D$.  Specifically, the binary bits
correspond to the $N$ bits of $\mathbf{x}^{[0]}$ and $M$ FIFO queues.
The floating-point numbers correspond to eligibility traces
($\alpha_{i,j,k}$ and $\gamma_{i,k}$ for $i,j\in[1,N]$, $k\in{\cal
  K}$, and $\ell\in{\cal L}$), the coefficients of the weight
($u_{i,j,k}$ and $v_{i,j,\ell}$ for $i,j\in[1,N]$, $k\in{\cal K}$, and
$\ell\in{\cal L}$), and the bias ($b_j$ for $j\in[1,N]$).

Each of the parameters of the DyBM-$\infty$ can be updated in a
distributed manner by the use of the learning rules from
(\ref{eq:learn_b})-(\ref{eq:learn_v}).  Observe that this distributed
update can be performed in constant time that is independent of $N$,
$D$, $|\cal K|$, and $|{\cal L}|$.

According to (\ref{eq:EXj}) and (\ref{eq:Ej}), the 
neuron $j$ is more likely to fire ($x_j^{[0]}=1$)
when (\rm i) $b_j$ is high, (\rm ii) $u_{i,j,k}$ and $\alpha_{i,j,k}$
are high, or (\rm iii) both conditions are met.  The learning rule
(\ref{eq:learn_b}) suggests that $b_j$ increases over time, if
the neuron $j$ fires more often than it is
expected from the latest values of the parameters of the
DyBM-$\infty$.  The learning rule (\ref{eq:learn_u}) suggests that
$u_{i,j,k}$ increases over time, if the neuron $j$
fires more often than it is expected, and the magnitude of
the changes in $u_{i,j,k}$ is proportional to the magnitude of
$\alpha_{i,j,k}$.  These implement long term potentiation.

According to (\ref{eq:EXj}) and (\ref{eq:Ej}), the neuron $j$ is less
likely to fire when (\rm i) $v_{i,j,\ell}$ and $\beta_{i,j,\ell}$ are
high, (\rm ii) $v_{j,i,\ell}$ and $\gamma_{i,\ell}$ are high, (\rm
iii) or both conditions are met.  The learning rule (\ref{eq:learn_v})
suggests that $v_{i,j,k}$ increases over time, if the neuron $j$
fires less often than it is expected, and the magnitude of the changes
in $v_{i,j,k}$ is proportional to the magnitude of $\beta_{i,j,\ell}$.
When $i$ and $j$ are exchanged, the learning rule (\ref{eq:learn_v})
suggests that $v_{i,j,k}$ increases over time, if the neuron $j$
fires less often than it is expected, and the magnitude of the changes
in $v_{i,j,k}$ is proportional to the magnitude of $\gamma_{i,\ell}$.
These implement long term depression.

Here, the terms in (\ref{eq:learn_b})-(\ref{eq:learn_v}) that involve
expected values (\ref{eq:EXj}) can be considered as a mechanism of
homeostatic plasticity \cite{lazar2009sorn} that keeps the firing
probability relatively constant.  This particular mechanism of
homeostatic plasticity does not appear to have been discussed with
STDP in the literature~\cite{lazar2009sorn,turrigiano2004homeostatic}.
We expect, however, that this formally derived mechanism of homeostatic
plasticity plays an essential role in stabilizing the learning of
artificial neural networks.  Without this homeostatic plasticity, the
values of the parameters can indeed diverge or fluctuate during
training.


\section{Conclusion}

Our work provides theoretical underpinnings on the postulates about
STDP.  Recall that the Hebb rule was first postulated in the middle of
the last century \cite{hebb1949organization} but had seen limited
success in engineering applications until more than 30 years later
when the Hopfield network \cite{hopfield1982neural} and the Boltzmann
machine \cite{hinton1983optimal} are used to provide theoretical
underpinnings.  In particular, a Hebbian rule was shown to increase
the likelihood of data with respect to the distribution associated
with the Boltzmann machine \cite{hinton1983optimal}.  STDP has been
postulated for biological neural networks and has been used for
artificial neural networks but in rather ad hoc ways.  Our work
establishes the relation between STDP and the Boltzmann machine for
the first time in a formal manner.

Specifically, we propose the DyBM as a stochastic model of
time-series.  The DyBM gives the conditional probability of the next
values of a multi-dimensional time-series given its historical values.
This conditional probability can depend on the whole history of
arbitrary length, so that the DyBM (specifically, DyBM-$\infty$) does
not have the limitation of a Markov model or a higher order Markov
model with a finite order.  The conditional probability given by the
DyBM-$\infty$ can thus be applied recursively to obtain the
probability of generating a particular time-series of arbitrary
length.

The DyBM-$\infty$ can be trained in a distributed manner (i.e., local
in space) with limited memory (i.e., local in time) when its
parameters have a proposed structure.  The learning rule is local in
space in that the parameters associated with a pair of the units in
the DyBM-$\infty$ can be updated by using only the information that is
available locally in those units.  The learning rule is local in time
in that it requires only a limited length of the history of a
time-series.  This training is guaranteed to converge as long as
the learning rate is set sufficiently small.

The DyBM-$\infty$ having the proposed structure (i.e., the homogeneous DyBM) can be considered as a
recurrent neural network, taking binary values, with memory units (or a
recurrent Boltzmann machine with memory).  Specifically, each neuron
stores eligibility traces and updates their values based on the spikes
that it generates and the spikes received from other neurons.  An axon
stores spikes that travel from a pre-synaptic neuron to a post-synaptic
neuron.  The synaptic weight is updated every moment, depending on the
spikes that are generated at that moment and the values of these
eligibility traces and the spikes stored in the axon.  This learning
rule exhibits various characteristics of STDP, including long term
potentiation and long term depression, which have been postulated and
observed empirically in biological neural networks.  The learning rule
also exhibits a form of homeostatic plasticity that is similar to those
studied for Bayesian spiking networks (e.g., \cite{HBN12}).  However,
the Bayesian spiking network is a mixture-of-expert model, which is a
particular type of a directed graphical model, while we study a
product-of-expert model, which is a particular type of an undirected
graphical model.

We expect that the theoretical underpinnings on STDP provided in this
paper will accelerate engineering applications of STDP.  In
particular, the prior work
\cite{hinton1999spiking,sutskever2007learning,sutskever2008recurrent,taylor2009factored}
has proposed various extensions of the Boltzmann machine to deal with
time-series data, but existing learning algorithms for these extended
Boltzmann machines involve approximations.  On the other hand, the
homogeneous DyBM can be considered as a recurrent Boltzmann machine with
memory, which naturally extends the Boltzmann machine (at the
equilibrium state) by taking into account the dynamics and by
incorporating the memory.  STDP is to the DyBM what the Hebb rule is to the Boltzmann machine.


\section*{Acknowledgements}

This research was supported by CREST, JST.

\bibliographystyle{abbrv}
\bibliography{sciencebib}

\end{document}